%% file: noisy_label_main.tex
\documentclass[10pt,twocolumn,letterpaper]{article}

\usepackage{cvpr}
\usepackage{times}
\usepackage{epsfig}
\usepackage{graphicx}
\usepackage{amsmath}
\usepackage{amssymb}

\usepackage{url}
\usepackage{tabularx}
\usepackage{caption}
\usepackage{subcaption}
\usepackage{booktabs}
\usepackage{wrapfig}
\usepackage{lipsum}
\usepackage{titling}
\usepackage{stfloats}

\usepackage[breaklinks=true,bookmarks=false]{hyperref}

\cvprfinalcopy % *** Uncomment this line for the final submission
\def\cvprPaperID{16} % *** Enter the CVPR Paper ID here

\ifcvprfinal\fi
\begin{document}

%%%%%%%%% TITLE
\title{Uncertainty Based Detection and Relabeling of Noisy Image Labels}

\date{}
%\author{Anonymous CVPR submission\\
%		\\
%            Paper ID 16}

\author{Jan M. K\"ohler\\
BCAI\\
%BCAI\thanks{Bosch Center for Artificial Intelligence. First name.last name@de.bosch.com}\\
%{\tt\small William.Beluch@de.bosch.com}
%%% For a paper whose authors are all at the same institution,
%%% omit the following lines up until the closing ``}''.
%%% Additional authors and addresses can be added with ``\and'',
%%% just like the second author.
%%% To save space, use either the email address or home page, not both
\and
Maximilian Autenrieth\\
BCAI, University of Ulm\\
\and
William H. Beluch\\
BCAI\thanks{Bosch Center for Artificial Intelligence. First name.last name@de.bosch.com}\\
%BCAI\\
%%{\tt\small Jan.Koehler@de.bosch.com}
}
\maketitle
%\thispagestyle{empty}

% soft way to not show page numbers
%\thispagestyle{empty}
%\vspace{1cm}

\input{abstract}

\input{introduction}

\input{related_work}

\input{methodology}

\input{experiments}

\input{conclusion}

\clearpage
\newpage
{\small
\bibliographystyle{ieee}
\bibliography{noisy_label}
}

%\newpage
%\input{appendix}

\end{document}

%% file: abstract.tex
\begin{abstract}
Deep neural networks (DNNs) are powerful tools in computer vision tasks. However, in many realistic scenarios label noise is prevalent in the training images, and overfitting to these noisy labels can significantly harm the generalization performance of DNNs. We propose a novel technique to identify data with noisy labels based on the different distributions of the predictive uncertainties from a DNN over the clean and noisy data. Additionally, the behavior of the uncertainty over the course of training helps to identify the network weights which best can be used to relabel the noisy labels. Data with noisy labels can therefore be cleaned in an iterative process. Our proposed method can be easily implemented, and shows promising performance on the task of noisy label detection on CIFAR-10 and CIFAR-100.  
\end{abstract}

%% file: introduction.tex
\section{Introduction}

In the last decade Deep neural networks (DNNs) have proven their predictive power in many supervised learning tasks with complex data patterns. Especially when the training set is large, representative, and correctly labeled, DNNs are the current state-of-the-art on several learning tasks. Unfortunately, the latter assumption does not hold in many realistic cases (e.g. medical imaging, crowd-sourced labeling), and DNNs have been shown to overfit on noisy labels, leading to poor generalization performance. For example, \cite{DBLP:journals/corr/ZhangBHRV16} shows that DNNs can easily fit randomly assigned labels on the training set, which leads to poor test performance.  Therefore, it is important to detect and correct for noisy labels in the training set. 

We propose an iterative label noise filtering process, based on the predictive uncertainty of the training images. Ensembles \cite{lakshminarayanan2017simple} and MC dropout \cite{gal2016dropout} are used to obtain uncertainty estimates for each image. We show that the uncertainties of the noisy images and the uncertainties of the clean images follow two different distributions, enabling the detection of potentially noisy labels. After the detection of the noisy labels, the detected image could be taken out of the training set, its weight on the loss could be decreased, or it could be relabeled through an oracle or any appropriate relabeling approach.

%% file: related_work.tex
\section{Related Work}

In the literature various approaches have been proposed to deal with label noise. 
\cite{ren2018learning} and \cite{jenni2018deep} utilize sample weights, derived from the performance of the network on a validation set, to reduce the influence of noisy labels. Other approaches exploit additional networks to assign sample weights by learning the structure of the label noise \cite{wang2018iterative,jiang2017mentornet,han2018co}. 
\cite{azadi2015auxiliary,natarajan2013learning,hendrycks2018noisy,zhang2018generalized,ma2018dimensionality,reed2014training} propose adjusted loss functions to diminish the influence of noisy labels during training.
\cite{brodley1999identifying} excludes potential noisy labels from training, with the disadvantage that information from the data is thrown away. 
Further approaches have recently been proposed to tackle the issue of label noise in image classification tasks, \cite{sukhbaatar2014learning,xiao2015learning,goldberger2017training,DBLP:journals/corr/Vahdat17,li2017learning,lee2018cleannet,yao2019deep,liu2016classification}; however, to the best of our knowledge, none of the proposed methods utilize model uncertainty to detect and filter out label noise in image classification tasks.  \cite{arazo2019beta} concurrently explore a very similar method to the one proposed in this paper by using a mixture of beta distributions over the training loss of noisy vs clean images.

%% file: methodology.tex
\section{Methodology}  \label{section_methodology}

In this section we explain our iterative, uncertainty-based, noise filtering process.
In Section \ref{section_experiments} the proposed method is then evaluated on two different noise patterns: Symmetric and Pair noise.  In the former, k$\%$ of the training labels are randomly flipped to another label ($k \in [20,40,60]$). In the latter,  $k\%$ of the labels are systematically flipped to the subsequent class label. Both noise patterns are very realistic in image classification tasks and currently discussed in the literature \cite{han2018co}.  The Pair setting is generally more difficult, especially when $k$ is greater than 50\%, as for a given real class, the class with the majority of labels is not the real class (e.g. a majority of the images with real label `dog' are labeled cat).  

\paragraph{Uncertainty acquisition} Let $y=g^W(x)$ be the output of a neural network with weights $W$ and input $x$, and $u=h(y|g^W,x)$ be the uncertainty of the model for its prediction $y$ given the input data $x$ and the model $g^W$. Since it has been shown that a single softmax score of one classifier does not serve well as an uncertainty measure \cite{gal2016dropout, gal2016uncertainty, hendrycks2017baseline},
we use three recent methods to easily obtain uncertainty estimates: Deep Ensembles \cite{lakshminarayanan2017simple} with $M$ members, Monte-Carlo dropout (MC-dropout \cite{gal2016uncertainty}), using $T$ forward passes, and a combination of both \cite{smith2018understanding}.

Having the predictions $y=g^{W_t}_m(x)$, with $t=1,\cdots, T$ forward passes and $m=1,\cdots, M$ classifiers, one needs a statistic $u(y)$ to quantify the uncertainty. The goal is to find an uncertainty statistic over the predictions $g^{W_t}_m(x)$  which depicts a high uncertainty value if $x$ has a noisy label and a low value if $x$ has a clean label. Many such statistics exist which have been successfully used in different settings \cite{beluch2018power, smith2018understanding, feinman2017detecting, gal2017deep, leibig2017leveraging, rawat2017adversarial}. 

We compare different statistics over the predictions $y=g^{W_t}_m(x)$ given by the multiple forward passes, including BALD \cite{houlsby2011bayesian}, Variation-Ratio, the standard deviation over the predictions $std(g^t_m(x))$ (averaged over all classes), and the mean over the predictions. For the mean we take the most likely of the $k$ classes of the softmax vector, \ie for brevity we denote $g^t_m(x) = \max_k g^{W_t}_m(x)$. \cite{hendrycks2017baseline} show that this maximum softmax probability is useful to distinguish between correctly and wrongly classified images.

\begin{figure}[t!]
	\begin{subfigure}[b]{0.49\linewidth}
		\includegraphics[width=0.95\linewidth]{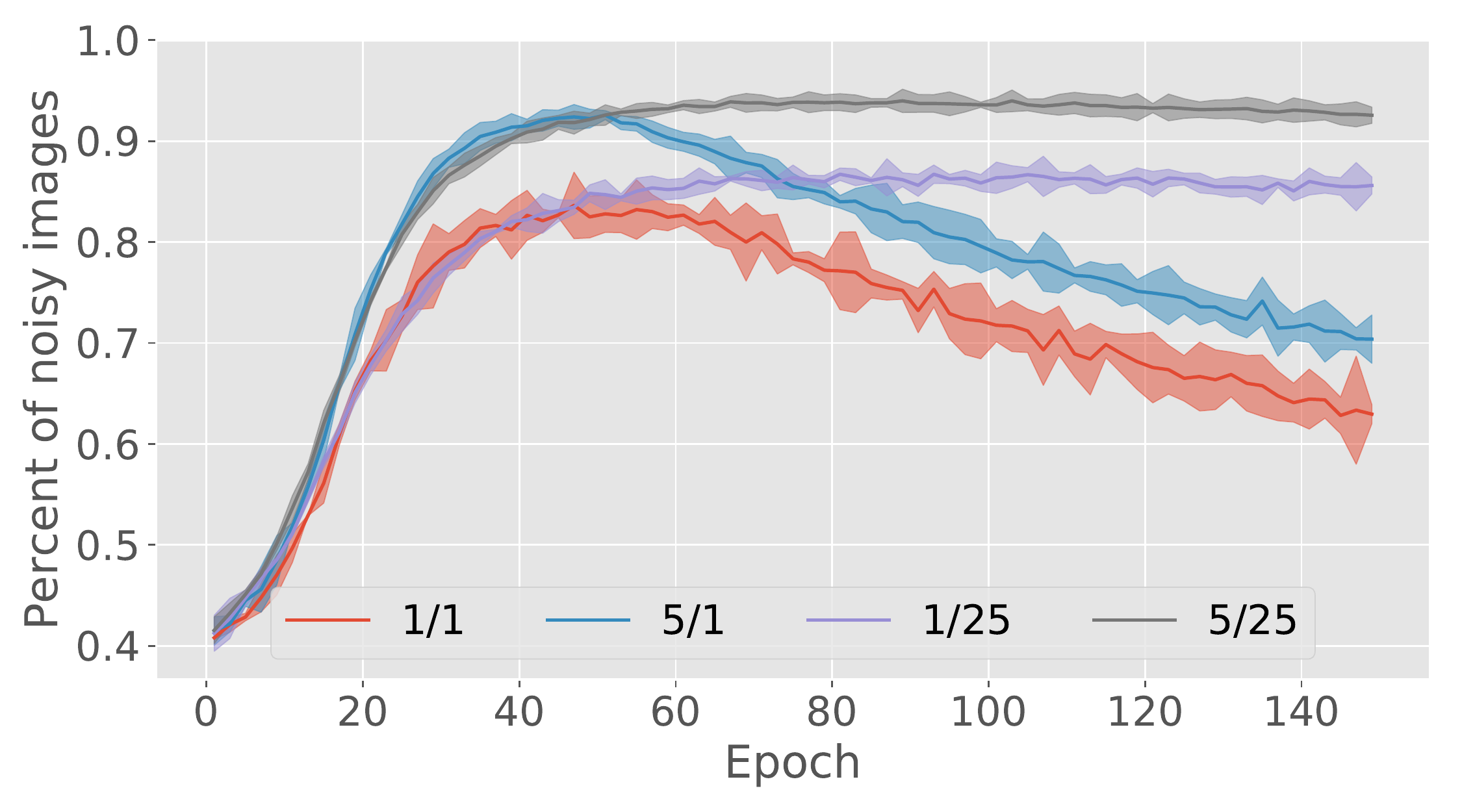}
		\caption{\footnotesize 40\%; S; 10\%; Softmax}
	\end{subfigure}
	\begin{subfigure}[b]{0.49\linewidth}
		\includegraphics[width=0.95\linewidth]{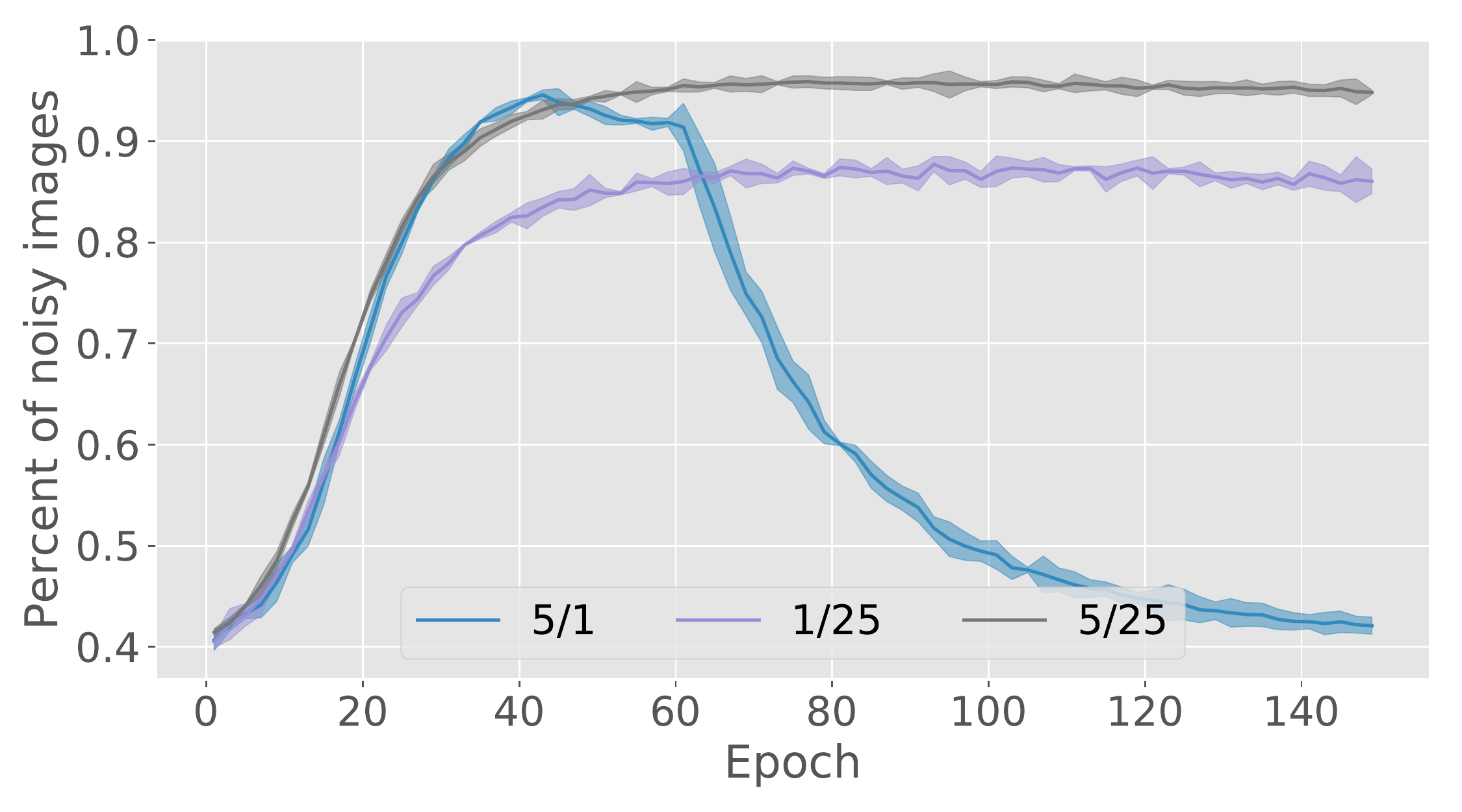}
		\caption{\footnotesize 40\%; S; 10\%; VR}
	\end{subfigure}
	
	\begin{subfigure}[b]{0.49\linewidth}
		\includegraphics[width=0.95\linewidth]{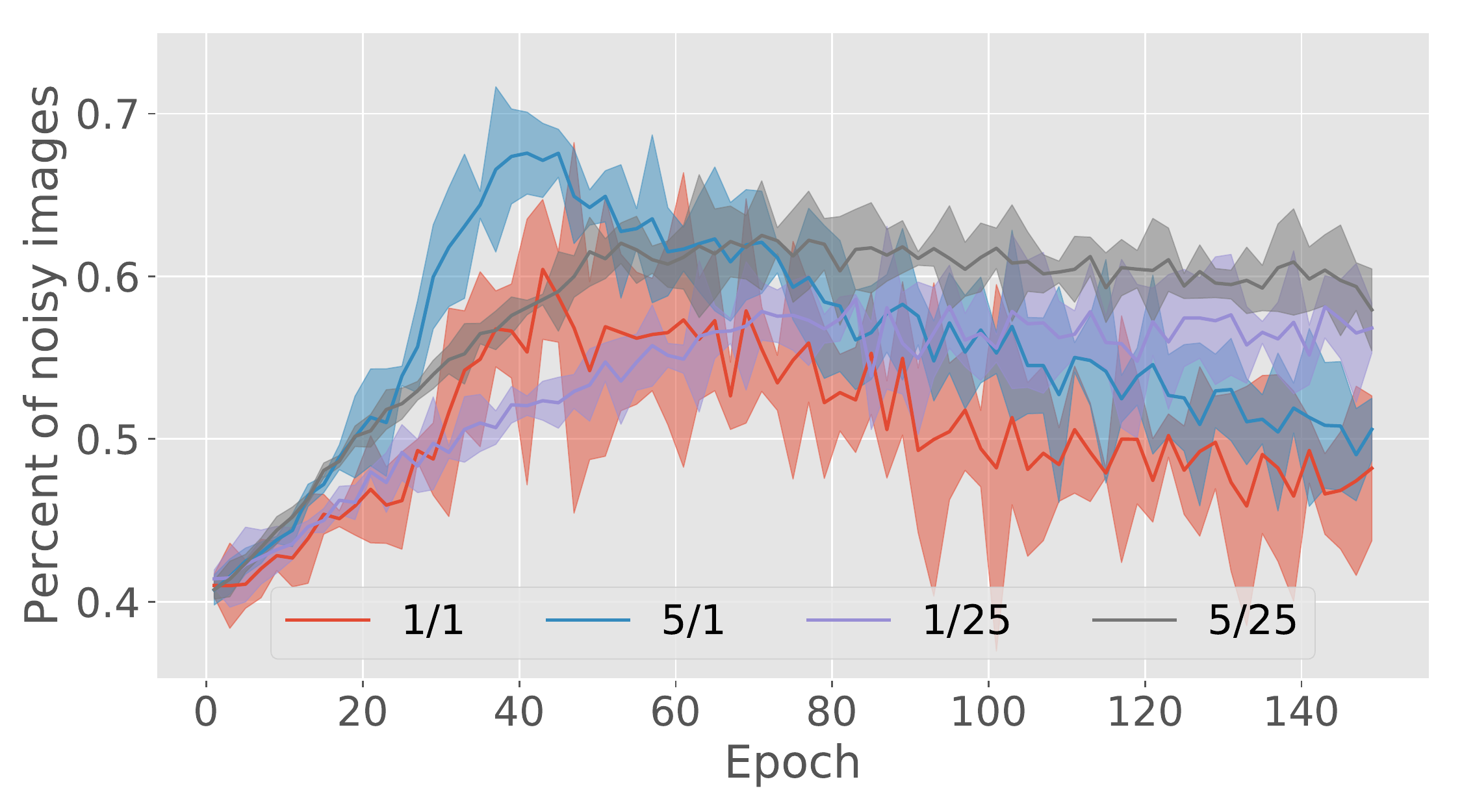}
		\caption{\footnotesize 40\%; P; 10\%; Softmax}
	\end{subfigure}
	\begin{subfigure}[b]{0.49\linewidth}
		\includegraphics[width=0.95\linewidth]{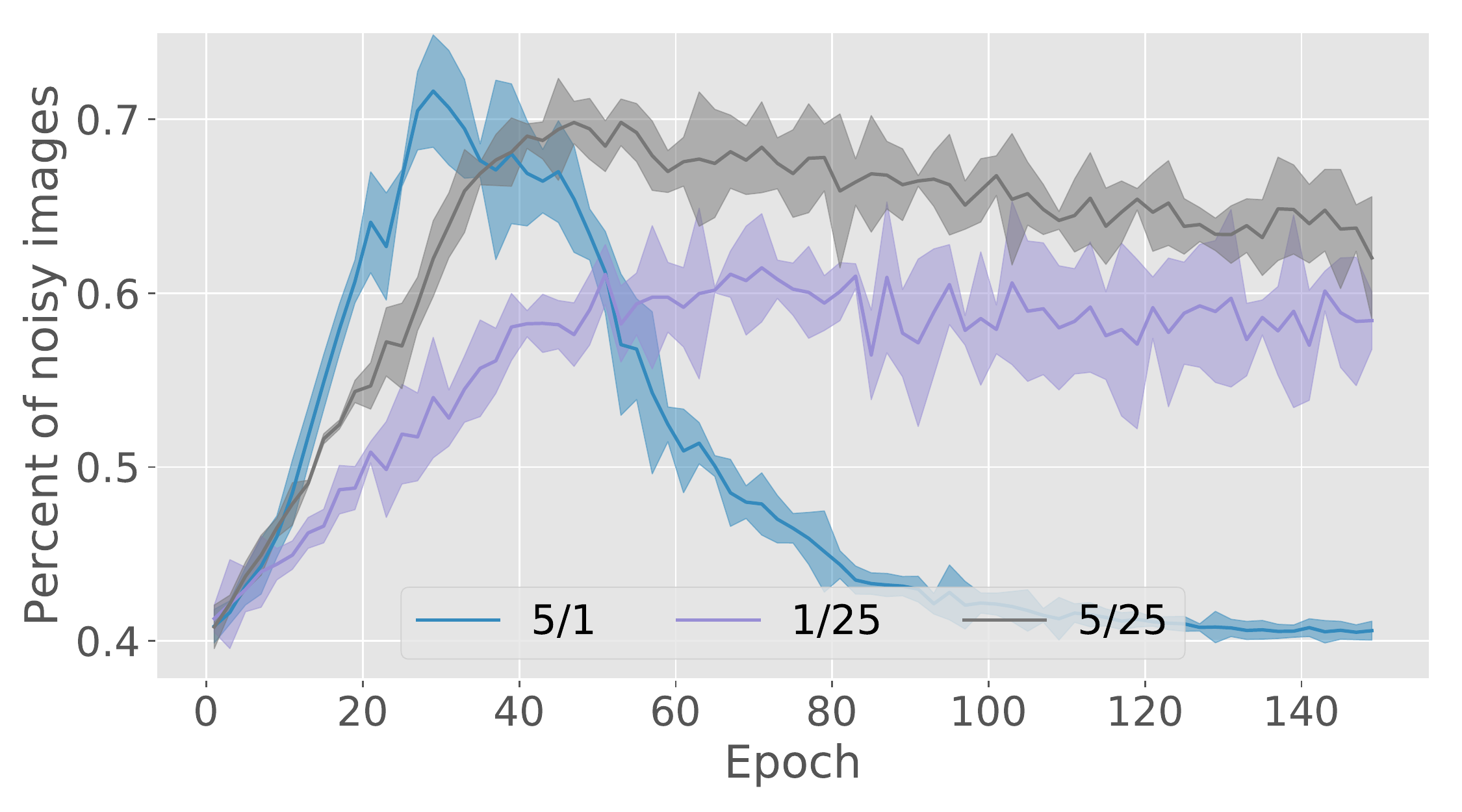}
		\caption{\footnotesize40\%; P; 10\%; VR}
	\end{subfigure}
	%\setbeamerfont{Figure 1}{size=\tiny}
	\caption{\footnotesize Ability to detect CIFAR-10 training images with noisy labels over the course of training.  The Y-axis is the percent of images in the subset of the X\% most uncertain images that have noisy labels.  The different colored-curves correspond to the model type used, and follow the format \# of classifiers / \# of stochastic forward passes per classifier.  The subcaptions follow the format: noise level; type of noise (P for pair, S for symmetric); uncertainty threshold (i.e. for 10\% on CIFAR-10, this would be the 5000 most uncertain images); uncertainty measure (VR=variation ratio).}
	\label{fig:compare_unc_measures}
\end{figure}

\paragraph{Noisy label detection}
We investigate the ability of the aforementioned uncertainty metrics to distinguish between noisy and clean labels. Our goal is to identify an epoch $T1$ in which the uncertainty of the noisy labels is significantly higher compared to the uncertainty of the clean labels.  Fig \ref{fig:compare_unc_measures} exhibits the ability of a given model type and a given uncertainty measure to detect noisy labels. At each training epoch, the training images with the highest $p\%$ uncertainty are selected, i.e. $X_p:= \{x: g_{m}^{W_t}(x) \geq x_{p}\}$. The proportion of images in this subset $(p=0.9)$ with a noisy label is plotted on the y-axis.  Note that this value should be scaled against the baseline noise level, i.e. if the baseline noise level is 40\%, the baseline ratio given by random sampling would be 40\%.

For both the Symmetric and Pair setting on CIFAR-10, the averaged softmax value of the predicted class and the Variation Ratio result in the highest selectivity (other uncertainty measures not shown).  Interestingly, the combination of MC-Dropout with an Ensemble is essential to obtain both a high selectivity and a robustness to long training times; using just MC-Dropout results in a lower peak ratio than an Ensemble, but when using an Ensemble with no stochastic forward passes the high selectivity lasts for only an epoch or two before rapidly tapering off.  

\begin{figure}[t!]
  \begin{subfigure}[b]{0.49\linewidth}
    \includegraphics[width=0.95\linewidth]{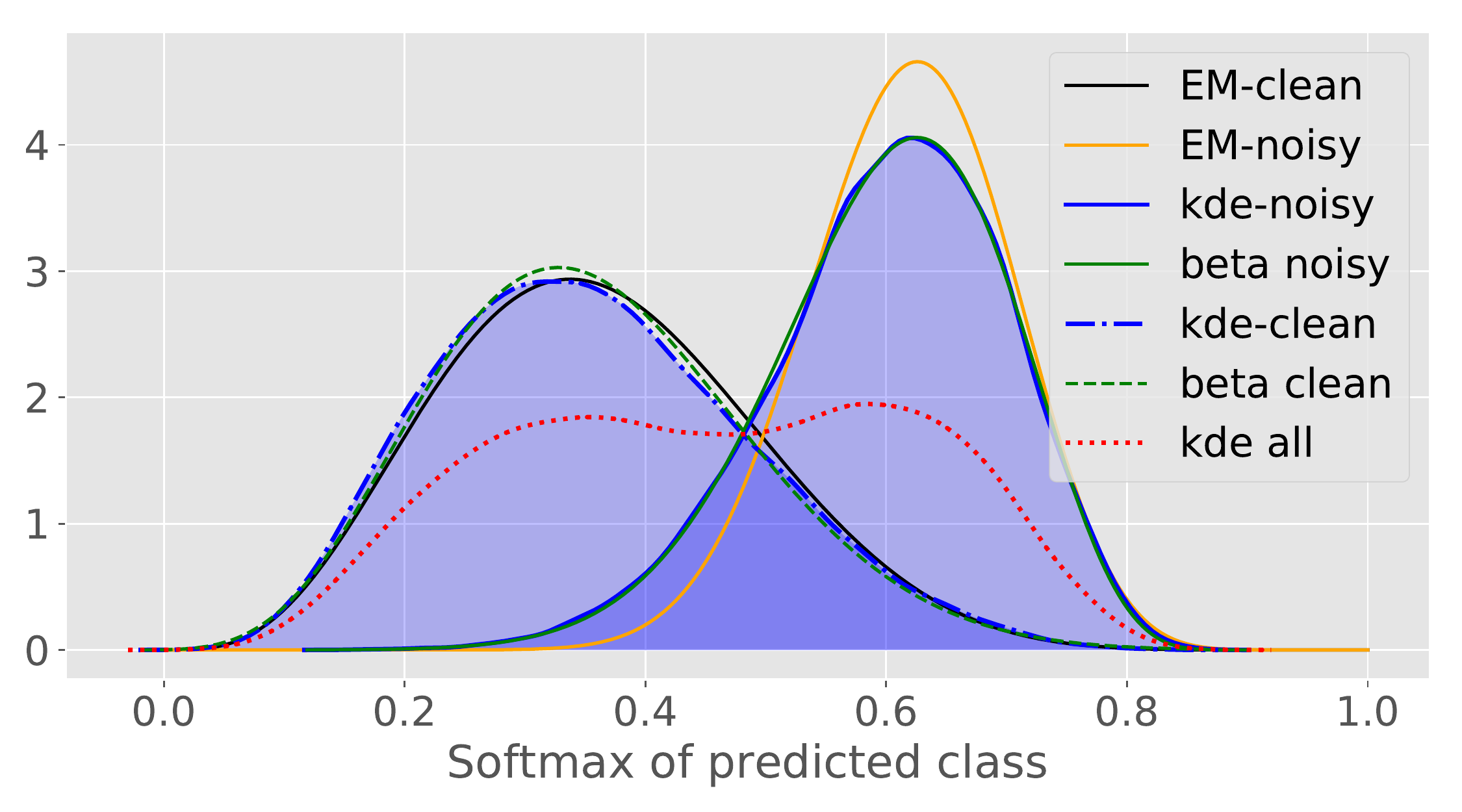}
    \caption{\footnotesize CIFAR-10}
  \end{subfigure}
 \begin{subfigure}[b]{0.49\linewidth}
    \includegraphics[width=0.95\linewidth]{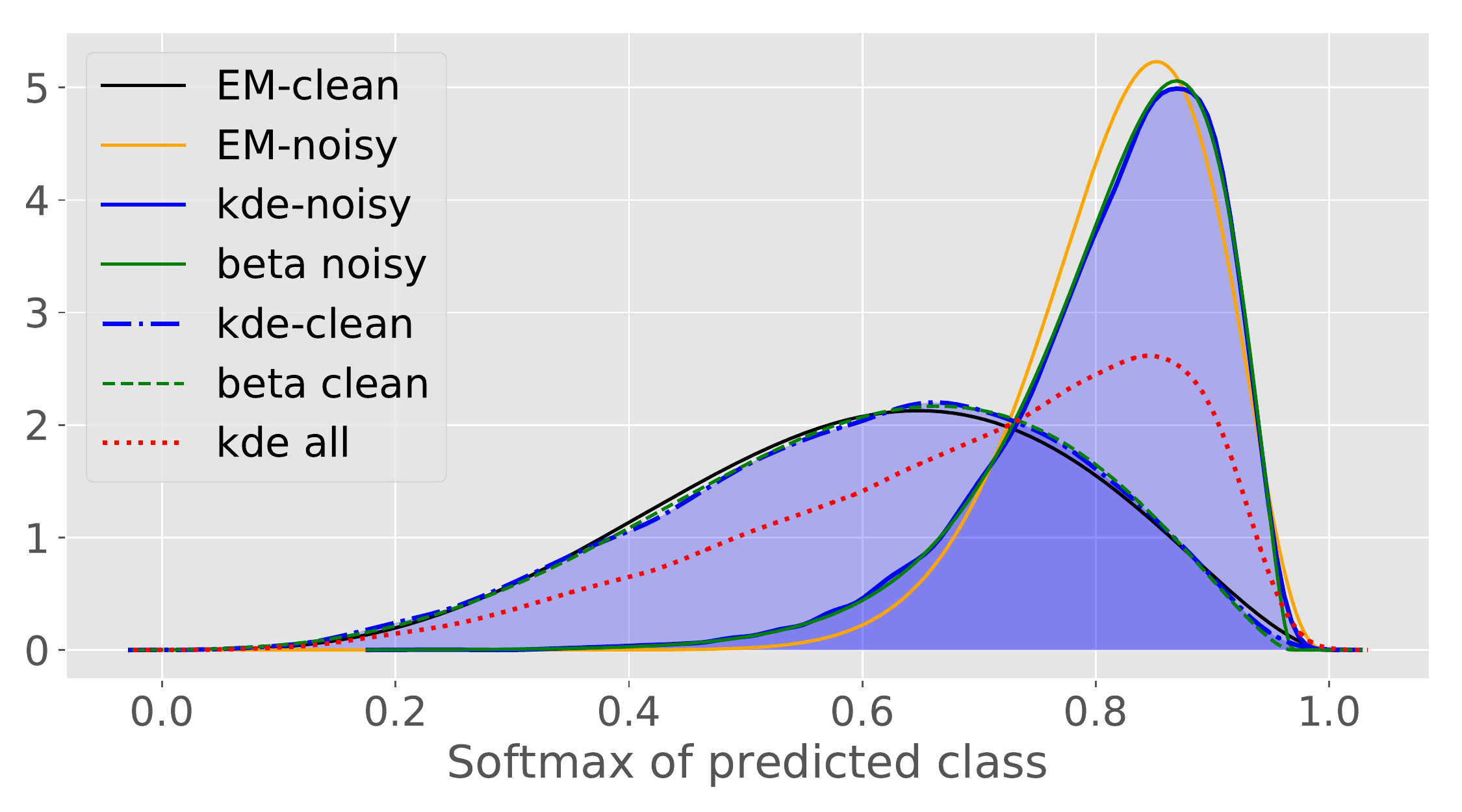}
    \caption{\footnotesize CIFAR-100}
  \end{subfigure}
  
 \caption{\footnotesize Expectation Maximization (EM) fits of two beta distributions (EM-clean / EM-noisy) to the averaged softmax values of images with clean and noisy labels. The overall distribution and its mixture components are plotted via kernel density estimations (kde all, kde-noisy, kde-clean) and their fit with a parametrized beta distribution (beta noisy, beta clean).}
  \label{fig:em}
\end{figure}

The optimal uncertainty threshold is highly dependent on many things, e.g.  the data set, noise type, noise level, and model architecture.  Within one experiment, as there are less noisy labels as the iterative process progresses, it makes sense to be more selective with the threshold. Instead of taking a fixed $p\%$ of most uncertain images, a possible extension is to model the distributions of the uncertainties of the noisy and clean images.  Empirically, a mixture of two beta distributions fits rather well to this task , and after using the Expectation Maximization (EM) algorithm \cite{dempster1977maximum} to fit the distributions (Fig \ref{fig:em}), the threshold can be chosen to be more selective, i.e. the number of clean images identified as noisy can be explicitly controlled for.    

\begin{figure}[t!]
	\begin{subfigure}[b]{0.49\linewidth}
		\includegraphics[width=0.95\linewidth]{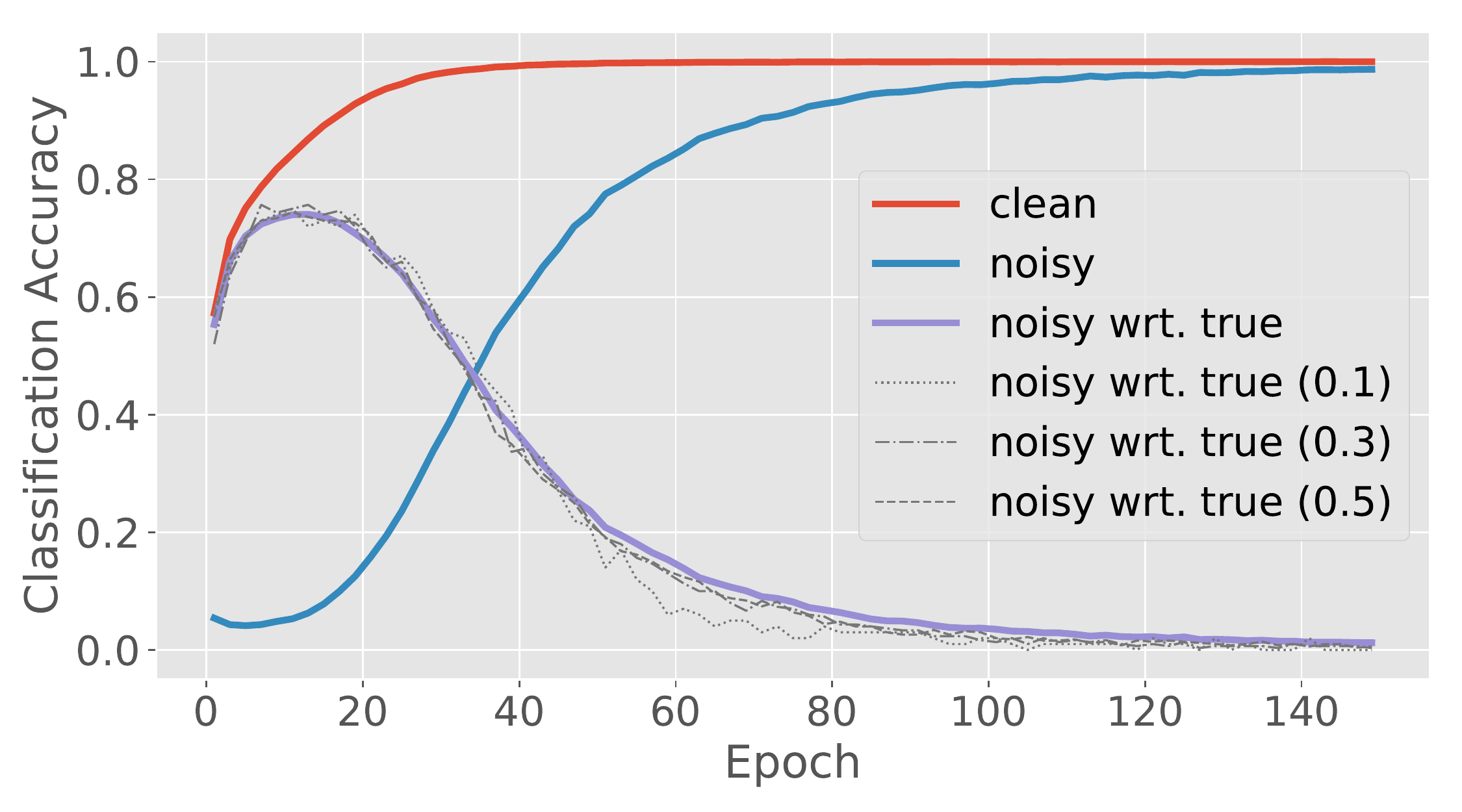}
		\caption{\footnotesize Accuracy of the noisy and clean CIFAR-10 training images over the course of training (40\% symmetric noise). Black curves correspond to training subsets with known noise.} 
	\end{subfigure}
	\begin{subfigure}[b]{0.49\linewidth}
		\includegraphics[width=0.95\linewidth]{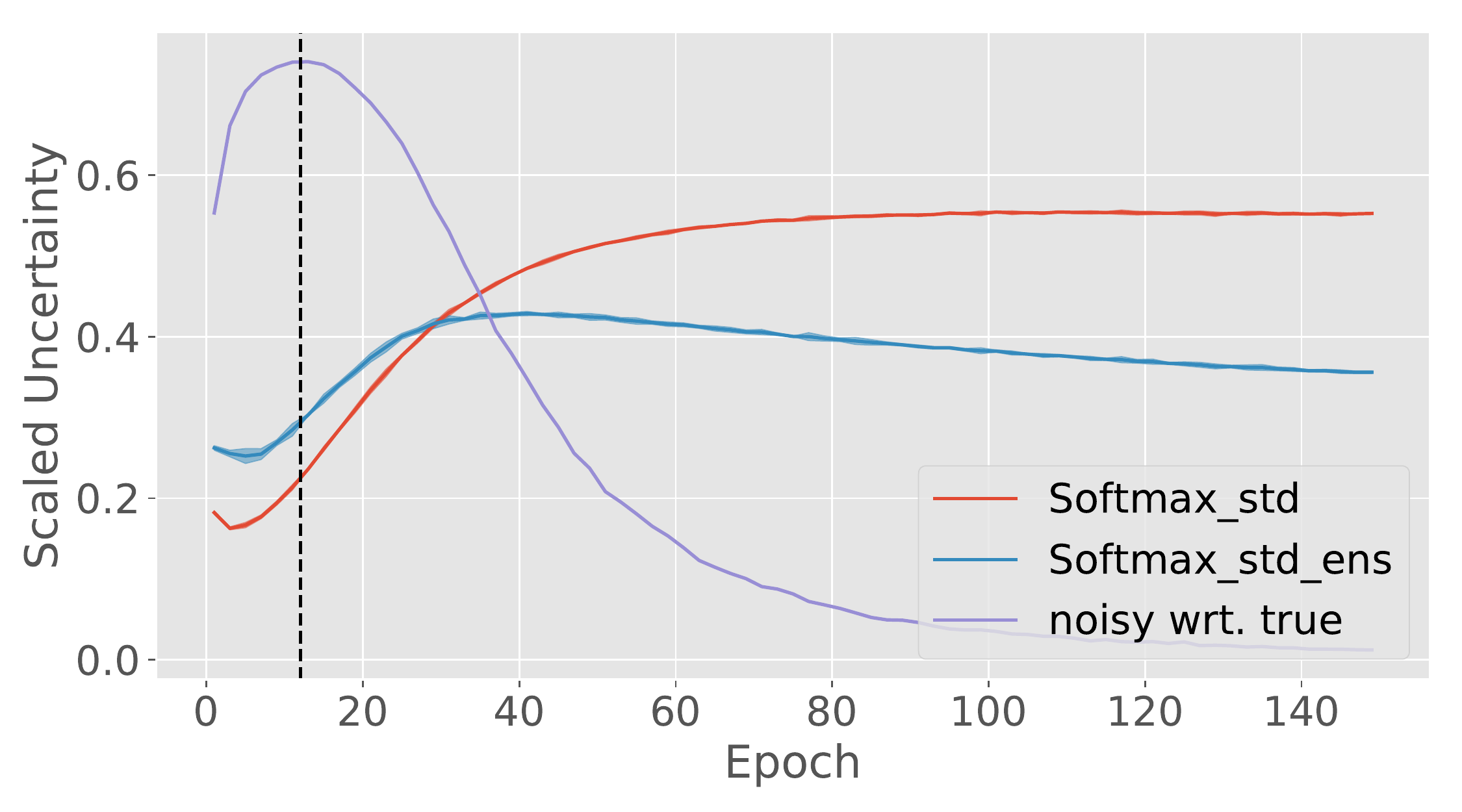}
		\caption{\footnotesize Mean uncertainty (std. dev. of softmax vectors) over all correctly classified CIFAR-10 training images over the course of training (40\% symmetric noise).}
	\end{subfigure}

	\caption{Relabeling of noisy images.}
	\label{fig:compare_relabel}
\end{figure}

\paragraph{Relabeling of noisy labels}
After noisy labels are detected, the goal is to relabel these images with the true label.  The simplest approach is to use the network's prediction for a noisy image before the network has overfit to the noisy labels.  Looking at Fig \ref{fig:compare_relabel}, this begins to happen at a relatively early phase, and ultimately, the network learns to predict the noisy labels perfectly.  However, around epoch 15 the network has learned enough from the clean images to correctly relabel 80\% of the noisy images (purple curve), before the overfitting to the noisy labels has begun.  

Of course this information is not available during training, as it requires knowledge of the true label for noisy images.  Thus two alternative approaches are proposed: the first uses a random subset of training images (e.g. 1000 images), in which the label noise is known (e.g. through expert relabeling; this is a common setting in the literature, e.g. \cite{ren2018learning}), and that the network is trained on.  Fig \ref{fig:compare_relabel}a shows that it is not required to know the true noise level in the data set; the curves behave similarly, and are sufficient proxies to identify at which epoch to use the predictions for relabeling. 

The second approach leverages the behavior of certain uncertainty estimates over the course of training, and requires no ground truth subset.  The std. dev. over the softmax outputs of the multiple forward passes, averaged over all training examples, first briefly sinks, before beginning to rise as the predicted class of the noisy labeled-images switches from the true label to the noisy label (Fig \ref{fig:compare_relabel}b).  This trend can be leveraged to roughly identify a good point to relabel using the predicted class.  Averaging the forward passes within one classifier and taking the std. dev. over the resulting 5 vectors (blue curve) provides a better heuristic for identifying the ideal relabel time than taking the standard deviation over all 125 total forward passes.

\vspace{-2mm}

%% file: experiments.tex
\section{Experimental results}  \label{section_experiments}

As an initial proof of concept experiment, we tackle the task of noisy label detection on CIFAR-10 and CIFAR-100, using the simple convolutional network described in \cite{beluch2018power}.  The noisy images are identified by taking the top 10\% of most uncertain images ($p=0.9$) (as in Fig\ref{fig:compare_unc_measures}a).  In Table \ref{u10results} the relabeling is based on the predicted softmax at an epoch determined by the criteria presented in Fig \ref{fig:compare_relabel}b. The results of 5 iterations of this process are shown, starting with 40\% symmetric noise.  The algorithm is able to reduce the number of noisy labels by almost half.  However, the accuracy of the trained networks do not rise (on a full clean data set this network achieves ~87\% accuracy).  Further analysis reveals that the images identified as noisy and correctly relabeled simply are not helpful in making the network generalize and be more robust to the label noise; the images that remain noisy, which are more difficult to relabel, are those important for increasing the classifier's performance. 

To highlight the effectiveness of the approach at detecting noisy labels, the experiment is repeated with oracle relabeling, in which all identified noisy images are given the correct true label (Table \ref{u10oracle}).  Now the accuracy of the network rises as the number of noisy labels drops.  At the end of the process, 16329 out of 20000 detected images were correctly identified as noisy and relabeled.  As expected, the precision of the detection drops as there are fewer noisy labels in the data set, yet at each iteration a fixed 10\% of images (i.e. 5000) with the highest uncertainty are selected. 

\begin{table}[t!]\
\footnotesize
\captionsetup{singlelinecheck=false}
\centering
\begin{tabular}{lllll} % .97\textwidth or XXX to devide equally table width
\toprule
  \textbf{Iter.} &  \textbf{Acc.} & \textbf{\# Noisy Images} & \textbf{Noise Prop.} & \textbf{Det. Prec.}  \\
\midrule
1  & 0.775  &  20000 &  0.400  &  0.917 \\
2  & 0.775  & 16803  &  0.336 &  0.852 \\  
3  & 0.775  & 13979  & 0.280  &  0.722 \\
4  & 0.773  & 11804  & 0.236  &  0.576 \\
5 & 0.767  & 10773  & 0.215  &  - \\
\bottomrule
\end{tabular}
\caption{\footnotesize Iterative relabeling based on predicted softmax on CIFAR-10.  Det.Prec = Detection Precision}
\label{u10results}
\end{table}

\begin{table}[t!]
\footnotesize
\captionsetup{singlelinecheck=false}
\centering
\begin{tabular}{lllll} % .97\textwidth or XXX to devide equally table width
\toprule
  \textbf{Iter.} &  \textbf{Acc.} & \textbf{\# Noisy Images} & \textbf{Noise Prop.} & \textbf{Det. Prec.}  \\
\midrule
1  & 0.773 (0.477)  &  20000 (20000) & 0.400 (0.400) & 0.943 (0.868) \\
2  & 0.796 (0.513)  & 15284 (15660) & 0.306  (0.313) & 0.902 (0.756)\\  
3  & 0.812 (0.535) & 10775 (11881) & 0.216  (0.237) & 0.796 (0.611)\\
4  & 0.826 (0.557) & 6797  (8825)&  0.136 (0.177) & 0.625 (0.451)\\
5 & 0.847  (0.572) & 3671  (6572) &  0.074 (0.131) & - \\

\bottomrule
\end{tabular}
\caption{\footnotesize Iterative relabeling with oracle relabeling on CIFAR-10 and CIFAR-100 (in parentheses). Det.Prec = Detection Precision}
\label{u10oracle}
\end{table}

%% file: conclusion.tex
\section{Conclusion}

We have shown that the predictive uncertainty given by a combination of an ensemble and MC-Dropout is very effective at identifying noisy labels under multiple noise settings and different datasets.  Further work is focused on improving both the detection and the relabeling.  For the former case, the beta-distribution fit will be further extended to increase the precision of detection at later stages in the algorithm.  Possible extensions to the relabeling include using a majority vote instead of the predicted softmax, or assigning multiple labels to an image that is difficult to relabel, with the idea that at least one of them is the correct one.  This can be done for a single model, or across model (i.e. each member of the ensemble gets a different label for an image). Finally, the methods will be tested on more data sets and network architectures, and compared to the state-of-the-art results from the literature.